\documentclass[11pt]{article}

\usepackage[final]{acl}

\usepackage{times}
\usepackage{latexsym}
\usepackage{booktabs}
\usepackage{tabularx}
\usepackage{siunitx}
\usepackage{multirow}
\usepackage{float}
\usepackage{kotex}

\usepackage[T1]{fontenc}

\usepackage[utf8]{inputenc}

\usepackage{microtype}

\usepackage{inconsolata}

\usepackage{graphicx}

\title{
Multi-View Attention Multiple-Instance Learning\\
Enhanced by LLM Reasoning for Cognitive Distortion Detection
}

\author{
\textbf{Jun Seo Kim}\textsuperscript{1*} \quad
\textbf{Hyemi Kim}\textsuperscript{2} \quad
\textbf{Woo Joo Oh}\textsuperscript{2} \quad
\textbf{Hongjin Cho}\textsuperscript{2} \\
\textbf{Hochul Lee}\textsuperscript{2} \quad
\textbf{Hye Hyeon Kim}\textsuperscript{3\dag} \\
\textsuperscript{1}Gachon University \quad
\textsuperscript{2}Korea Telecom Research \quad
\textsuperscript{3}Yonsei University \\
\texttt{kma80kjs@gachon.ac.kr} \\
\texttt{\{mika.kim, woojoo.oh, as.df, hochul.lee\}@kt.com} \\
\texttt{hye\_hyeon@yonsei.ac.kr}
}

\begin{document}
\maketitle
\renewcommand{\thefootnote}{}
\footnotetext{\textsuperscript{*}First author. \textsuperscript{\dag}Corresponding author.}
\setcounter{footnote}{0}
\renewcommand{\thefootnote}{\arabic{footnote}}

\begin{abstract}
Cognitive distortions have been closely linked to mental health disorders, yet their automatic detection remains challenging due to contextual ambiguity, co-occurrence, and semantic overlap. We propose a novel framework that combines Large Language Models (LLMs) with a Multiple-Instance Learning (MIL) architecture to enhance interpretability and expression-level reasoning. Each utterance is decomposed into Emotion, Logic, and Behavior (ELB) components, which are processed by LLMs to infer multiple distortion instances, each with a predicted type, expression, and model-assigned salience score. These instances are integrated via a Multi-View Gated Attention mechanism for final classification. Experiments on Korean (KoACD) and English (Therapist QA) datasets demonstrate that incorporating ELB and LLM-inferred salience scores improves classification performance, especially for distortions with high interpretive ambiguity. Our results suggest a psychologically grounded and generalizable approach for fine-grained reasoning in mental health NLP. The dataset and implementation details are publicly accessible\footnote{\url{https://github.com/cocoboldongle/MVACD}}.

\end{abstract}

\begin{table*}[t]
\centering
\small
\begin{tabularx}{\textwidth}{l X X}
\toprule
\textbf{Cognitive Distortion Type} & \textbf{Definition} & \textbf{Examples} \\
\midrule
All-or-Nothing Thinking & Viewing situations in only two categories (e.g., perfect or failure) instead of on a spectrum. & “If I fail this test, I’m a total failure.” \\

Jumping to Conclusions & Predicting negative outcomes without evidence. & “She didn’t text back. She must be mad at me.” \\

Personalization & Blaming yourself for events outside your control or assuming excessive responsibility. & “My friend looks sad, maybe I did something wrong.” \\
\bottomrule
\end{tabularx}
\caption{Definitions and examples of selected cognitive distortion types adapted from \citep{kim2025}.}
\label{tab:cog_distortion}
\end{table*}

\section{Introduction}

Mental illness is a widespread global health concern. About half of the global population experience mental illness during their lifetimes, and one in eight is affected at any given time \citep{mcgrath2023, who2022}. Mental health conditions such as anxiety, depression, and emotional expression difficulties are significantly associated with cognitive distortions, indicating their role in both the formation and persistence of emotional distress \citep{mercan2023}.

Cognitive distortions refer to systematic errors in thinking that occur when individuals perceive and interpret external information, leading to a negative conclusion that does not correspond to reality \citep{beck1979}. Table~\ref{tab:cog_distortion} summarizes the definitions and illustrative examples of a subset of cognitive distortion types considered in this study.

Influenced by internal factors such as emotions and beliefs, these distortions are expressed through language or automatic thoughts, reinforcing emotional distress and maladaptive behavior \citep{strohmeier2016}. Cognitive distortions have played a central role in the onset and maintenance of various mental illnesses; therefore, identifying these patterns is often considered an important component of the therapeutic process \citep{morrison2015, kaplan2017}.

Recently, there have been active attempts to automatically detect cognitive distortions in mental health-related texts by utilizing the advanced language comprehension and reasoning capabilities of large language models (LLMs) \citep{chen2023, qi2023}. However, most existing studies have treated utterances as single, unstructured inputs, returning predictions for the entire text without considering the internal psychological structure of each utterance. In particular, they overlook the fact that different cognitive distortions may arise from distinct aspects of an utterance—such as emotion, logic, or behavior—and that these components interact to shape distorted thinking. As a result, the interplay of such psychological factors often remains underrepresented, limiting interpretability and the granularity of model inference. Moreover, multiple cognitive distortions often occur together in a single utterance, and the semantic similarity between types can lead to differences in interpretation among experts and difficulties in establishing a gold standard \citep{suputra2023}.

To address these challenges, we propose a new cognitive distortion detection framework inspired by the Multiple-Instance Learning (MIL) structure \citep{dietterich1997}, in which an utterance is defined as a bag, and each of the multiple cognitive distortion expressions inferred by an LLM is considered an instance to make a final decision. Each instance includes the predicted distortion type, its associated sentence, and a salience score assigned by the LLM, which was incorporated into the final prediction through weighted aggregation within the MIL structure.

In addition, we designed this study to enable more precise and interpretable cognitive distortion reasoning by decomposing each utterance into three psychologically grounded components—Emotion, Logic, and Behavior (ELB)—and inputting them into the LLM along with the original text. This approach moves beyond previous methods that relied solely on a single-text input, enabling more precise and interpretable prediction of complex, overlapping cognitive distortions in real-world utterances.

Our primary contributions are summarized as follows:
\begin{enumerate}
    \item We obtained high-quality labels for 10 cognitive distortion types through expert review by 10 psychologists.
    \item We structured each utterance into three psychologically grounded components (ELB) and incorporated this information into the LLM input to support more informed and context-aware inference.
    \item We proposed the MIL-based framework that treats each LLM-inferred cognitive distortion as an instance, integrating both the predicted types and LLM-assigned salience score into a unified classification model.
\end{enumerate}

\section{Related Work}

\subsection{Cognitive Distortions Detection}

Early studies treated utterances as single units and applied binary or multi-class classification using Linguistic Inquiry and Word Count features and models such as logistic regression or SVMs \citep{simms2017, shreevastava2021}. While effective in binary settings, these approaches have often struggled with label imbalance and semantic overlap in multi-class scenarios.

To capture co-occurring distortions, later work introduced multi-label classification \citep{ding2022, shickel2020, elsharawi2024}, along with data augmentation and domain-adaptive language models. More recently, standardized re-annotation and teacher-student multi-task learning have been employed to address the lack of generalization across diverse datasets \citep{qi2025}. However, even in these recent studies, utterances are still processed holistically, without structural decomposition.

Some studies incorporated conversational context, modeling multi-turn interactions to improve continuity and prediction \citep{lybarger2022, tauscher2023}. Yet, these models also lack expression-level inference and focus primarily on dialogue flow.

More recently, large LLMs have been applied to cognitive distortion detection. The Diagnosis of Thought (DoT) framework introduced a structured prompting approach to improve interpretability \citep{chen2023}. Another study explored zero-shot and few-shot prompting for distortion classification without supervised training \citep{qi2023}.

Despite recent progress, prior work has not yet modeled cognitive distortions at the expression level or incorporate the co-occurrence of multiple distortions within an utterance into prediction. To address this, we propose a framework that decomposes utterances into the final ELB components and infers distortions as instances within a Multiple-Instance Learning setup for more interpretable and fine-grained classification.

\subsection{Multiple-Instance Learning in NLP}

Multiple-Instance Learning (MIL) is a weakly supervised framework in which multiple instances are grouped into a single bag and a prediction is made at the bag level. Originally proposed for drug activity prediction in bioinformatics \citep{dietterich1997}, MIL was later applied to computer vision tasks such as natural scene classification \citep{maron1998}, demonstrating its flexibility in learning from partially labeled data.

In NLP, MIL has been applied to tasks such as document classification, sentiment analysis, and misinformation detection. Early approaches employed machine learning models such as mi-SVM, MILBoost, and other instance-level classifiers adapted to weakly supervised settings \citep{andrews2002, zhang2008, jorgensen2008}.

Later work extended MIL with deep architectures to infer sentence-level sentiment from document-level labels. Approaches such as manifold regularization \citep{kotzias2014} and weighted instance modeling \citep{pappas2014} have improved both interpretability and prediction accuracy.

More recent studies have further enhanced MIL frameworks by integrating contextualized embeddings and attention mechanisms. For example, attention-based MIL was applied to fake news detection with improvements in precision and interpretability \citep{karaoglan2024}, and mutual-attention models were used to address bag-instance mismatch in hate speech classification \citep{liu2022}.

However, despite these advances, most MIL-based approaches in NLP define instances at the sentence or paragraph level, without leveraging finer-grained semantic representations. Incorporating LLM-inferred expressions and their types and salience scores into MIL has not yet been explored. We address this gap by proposing a model that treats each LLM-generated unit as an instance and integrates its label and salience score into attention-based bag-level classification.

\section{Dataset}

\begin{table}[t]
\renewcommand{\arraystretch}{1.2}
\centering
\small
\begin{tabularx}{\linewidth}{l >{\raggedleft\arraybackslash}X}
\toprule
\textbf{Set} & \textbf{Utterances (\%)} \\
\midrule
Train       & 3,608 (80\%) \\
Validation  & 451 (10\%) \\
Test        & 451 (10\%) \\
\midrule
\textbf{Cognitive Distortion Type} & \textbf{Count (\%)} \\
\midrule
Labeling                    & 478 (10.6\%) \\
Negative Filtering          & 470 (10.4\%) \\
All-or-Nothing Thinking     & 464 (10.3\%) \\
Emotional Reasoning         & 458 (10.2\%) \\
Personalization             & 459 (10.2\%) \\
Overgeneralization          & 452 (10.0\%) \\
Discounting the Positive    & 451 (10.0\%) \\
Jumping to Conclusions      & 431 (9.6\%) \\
Magnification and Minimization & 432 (9.6\%) \\
Should Statements           & 415 (9.2\%) \\
\midrule
\textbf{Total} & \textbf{4,510 (100.0\%)} \\
\bottomrule
\end{tabularx}
\caption{Statistics of the KoACD dataset.}
\label{tab:koacd}
\end{table}

\begin{table}[t!]
\renewcommand{\arraystretch}{1.2}
\centering
\small
\begin{tabularx}{\linewidth}{l >{\raggedleft\arraybackslash}X}
\toprule
\textbf{Set} & \textbf{Utterances (\%)} \\
\midrule
Train       & 1,277 (80\%) \\
Validation  & 159 (10\%) \\
Test        & 161 (10\%) \\
\midrule
\textbf{Cognitive Distortion Type} & \textbf{Count (\%)} \\
\midrule
Mind Reading            & 239 (15.0\%) \\
Overgeneralization      & 239 (15.0\%) \\
Magnification           & 195 (12.2\%) \\
Labeling                & 165 (10.3\%) \\
Personalization         & 153 (9.6\%) \\
Fortune-telling         & 143 (9.0\%) \\
Emotional reasoning     & 134 (8.4\%) \\
Mental filter           & 122 (7.6\%) \\
Should statements       & 107 (6.7\%) \\
All-or-nothing thinking & 100 (6.3\%) \\
\midrule
\textbf{Total} & \textbf{1,597 (100.0\%)} \\
\bottomrule
\end{tabularx}
\caption{Statistics of the Therapist QA dataset.}
\label{tab:therapist}
\end{table}

\setcounter{figure}{1}  
\begin{figure*}[t]
\centering
\includegraphics[width=\textwidth]{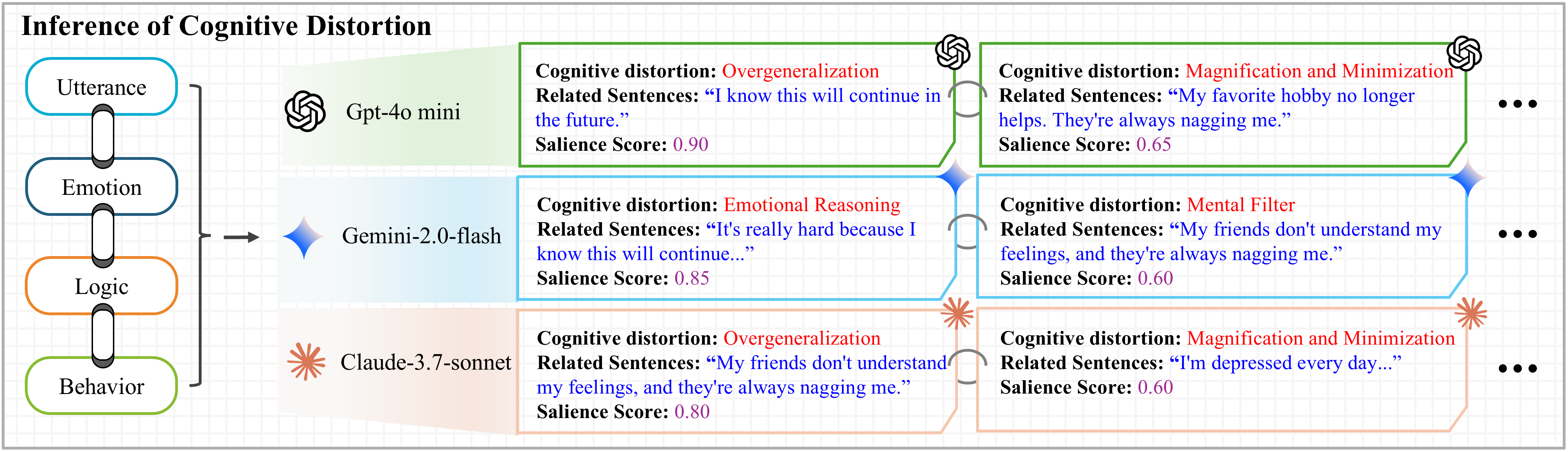}
\caption{LLM-based Inference of Cognitive Distortion Instances from ELB-Structured Utterances.}
\label{figure2}
\end{figure*}

\setcounter{figure}{0}  

\begin{figure}[t]
\centering
\includegraphics[width=\linewidth]{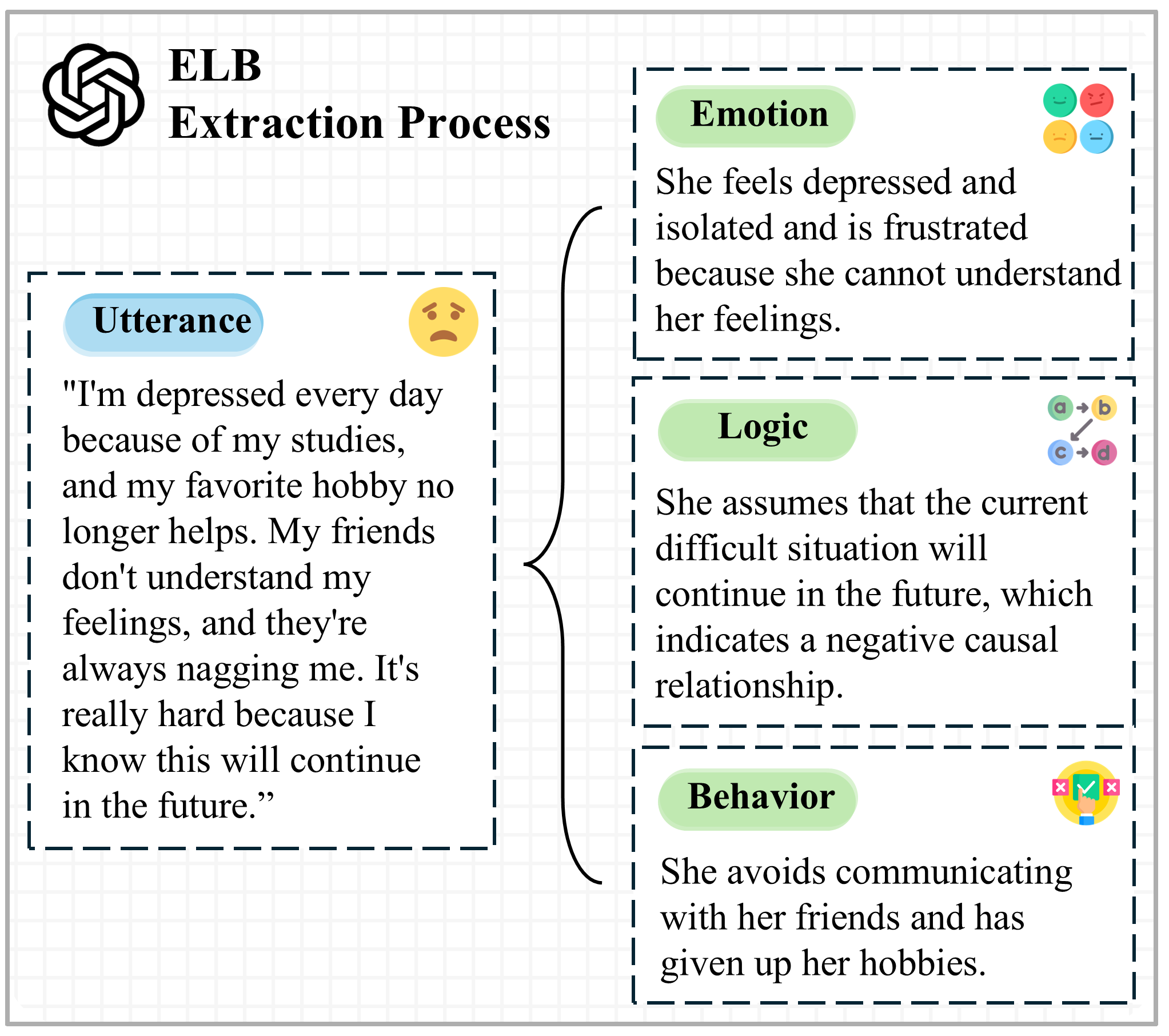}
\caption{ELB-Based Psychological Decomposition of an Utterance.}
\label{figure1}
\end{figure}

\paragraph{KoACD} The Korean Adolescent Cognitive Distortion dataset was derived from counseling texts collected from the NAVER Knowledge iN platform \citep{kim2025}. For this study, we sampled 5,000 utterances (500 per distortion type) and conducted expert validation with 10 Korean psychologists with master's degrees and over five years of experience. Each utterance was reviewed by a pair of experts to cross-check the original labels, and after removing disagreements, 4,510 utterances with a single validated distortion label were retained. The data were split into training, validation, and test sets in an 8:1:1 ratio, and the label distribution is shown in Table~\ref{tab:koacd}.

\paragraph{Therapist QA Dataset} This dataset consists of 1,597 expert-annotated English utterances from asynchronous patient–therapist logs available on the Kaggle platform \citep{shreevastava2021}. While the original data may contain up to two labels per utterance, we used the primary label for each instance. The dataset covers 10 distortion types and was split 8:1:1 to benchmark cross-linguistic generalization, as shown in Table~\ref{tab:therapist}. Both datasets used in this study are publicly available.

\section{Emotion--Logic--Behavior Extraction}

To better capture the psychological context of each utterance, we decompose it into three components---Emotion, Logic, and Behavior (ELB). This decomposition draws on the CBT cognitive triangle \citep{beck1979}, with ``Logic'' used in place of ``Thought'' to emphasize its reasoning-oriented nature in cognitive distortions. This structured representation is designed to improve both the interpretability and granularity of cognitive distortion inference by making explicit the latent psychological dimensions, which are often entangled in unstructured text.

As illustrated in Figure~\ref{figure1}, the ELB components are extracted using a zero-shot prompting strategy based on GPT-4 \citep{openai2023}, which is guided to independently generate each of the three elements for every utterance. These extracted components, combined with the original text, serve as enriched input to the downstream LLM-based inference process, enabling more context-aware and psychologically grounded predictions. The LLM hyperparameters and prompt templates are listed in Table~\ref{table11} in Appendix~\ref{AppendixB} and Table~\ref{table17} in Appendix~\ref{AppendixF}, respectively. To ensure reliable generation quality, we configure the LLM generation parameters to balance interpretive diversity with output stability and to accommodate variations in bag length. These settings are informed by empirical observations during development and aim to produce consistent yet sufficiently expressive outputs.

\section{LLM-Based Inference of Cognitive Distortion Instances}

To infer multiple cognitive distortion instances per utterance, we employ three LLMs—OpenAI GPT-4o \citep{openai2023}, Google Gemini 2.0 Flash \citep{deepmind2024}, and Anthropic Claude 3.7 Sonnet \citep{anthropic2025}. Each model independently processes the same utterance, using the pre-extracted ELB components as input to infer cognitive distortions. Each LLM infers a set of instances consisting of a predicted distortion type, a corresponding text segment, and an LLM-assigned salience score, as illustrated in Figure~\ref{figure2}. Due to semantic ambiguity, a single sentence can map to multiple distortion types, either simultaneously or through varying interpretations. To evaluate the contribution of ELB information, we also perform inference using only the original utterances. The LLM hyperparameters are listed in Table~\ref{table11} in Appendix~\ref{AppendixB}, and the prompt templates are presented in Tables~\ref{table18} and~\ref{table19} in Appendix~\ref{AppendixF}.

\section{Experiments}

\begin{table}[t]
\centering
\small
\renewcommand{\arraystretch}{1.2}
\begin{tabularx}{\linewidth}{l >{\raggedleft\arraybackslash}X}
\toprule
\textbf{Instance Statistics} & \textbf{Value} \\
\midrule
Total Instances & 52,201 \\
Min Instances per Bag & 5 \\
Max Instances per Bag & 20 \\
Avg. Instances per Bag & 11.57 \\
\midrule
\textbf{Cognitive Distortion Type} & \textbf{\# Instances (\%)} \\
\midrule
Jumping to Conclusions & 10,154 (19.5\%) \\
Overgeneralization & 9,942 (19.1\%) \\
Negative Filtering & 6,130 (11.7\%) \\
Emotional Reasoning & 6,095 (11.7\%) \\
Personalization & 5,531 (10.6\%) \\
All-or-Nothing Thinking & 4,739 (9.1\%) \\
Labeling & 3,580 (6.9\%) \\
Magnification and Minimization & 2,317 (4.4\%) \\
Should Statements & 2,190 (4.2\%) \\
Discounting the Positive & 1,523 (2.9\%) \\
\bottomrule
\end{tabularx}
\caption{Instance Distribution in KoACD (with ELB).}
\label{table4}
\end{table}

\begin{table}[t!]
\centering
\small
\renewcommand{\arraystretch}{1.2}
\begin{tabularx}{\linewidth}{l >{\raggedleft\arraybackslash}X}
\toprule
\textbf{Instance Statistics} & \textbf{Value} \\
\midrule
Total Instances & 13,967 \\
Min Instances per Bag & 4 \\
Max Instances per Bag & 24 \\
Avg. Instances per Bag & 8.75 \\
\midrule
\textbf{Cognitive Distortion Type} & \textbf{\# Instances (\%)} \\
\midrule
Emotional reasoning & 2,999 (21.5\%) \\
Overgeneralization & 2,271 (16.3\%) \\
Fortune-telling & 1,628 (11.7\%) \\
Mind Reading & 1,592 (11.4\%) \\
Magnification & 1,245 (8.9\%) \\
All-or-nothing thinking & 1,012 (7.2\%) \\
Personalization & 920 (6.6\%) \\
Mental filter & 899 (6.4\%) \\
Should statements & 783 (5.6\%) \\
Labeling & 618 (4.4\%) \\
\bottomrule
\end{tabularx}
\caption{Instance Distribution in Therapist QA (with ELB).}
\label{table5}
\end{table}

\subsection{MIL Setup and Bag Construction}

This study formulates cognitive distortion classification using a MIL framework, adopting an instance-based representation to effectively capture multiple distortions within a single utterance. To this end, we aggregate cognitive distortion representations generated by three LLMs. Each LLM independently infers multiple cognitive distortion representations for a given utterance $B$, with each instance $x_i$ defined by the predicted distortion type $\textit{type}_i$, the associated expression text $\textit{text}_i$, and a normalized salience score $s_i$ assigned by the LLM. The entire utterance $B$ is represented as a bag comprising a set of these instances. The overall construction is defined as follows:

\begin{equation}
B = \{x_i = (\textit{type}_i, \textit{text}_i, s_i )\}_{i=1}^{N}
\label{eq:bag}
\end{equation}

where $N$ is the total number of instances generated by the model. Each instance's salience score is normalized to reflect its relative importance within the bag. The normalization process is defined as follows:

\begin{equation}
\hat{p}_i = \frac{s_i}{\sum_{j=1}^{N} s_j}
\label{eq:norm}
\end{equation}

The resulting bag serves as input to the MIL model, with instance-level salience score values used as weighting signals within the attention-based integration structure. Table~\ref{table4} and Table~\ref{table5} present instance-level summary statistics for the KoACD and Therapist QA datasets, respectively, under the ELB-based inference setting. These tables report the number of instances per cognitive distortion type, as well as the average and maximum number of instances per bag. For reference, statistics obtained without incorporating ELB information are provided in Table~\ref{table9} and Table~\ref{table10} in Appendix~\ref{AppendixA}.

\subsection{Embedding Representation}
\setcounter{figure}{2}

\begin{figure*}[t]
\centering
\includegraphics[width=\textwidth]{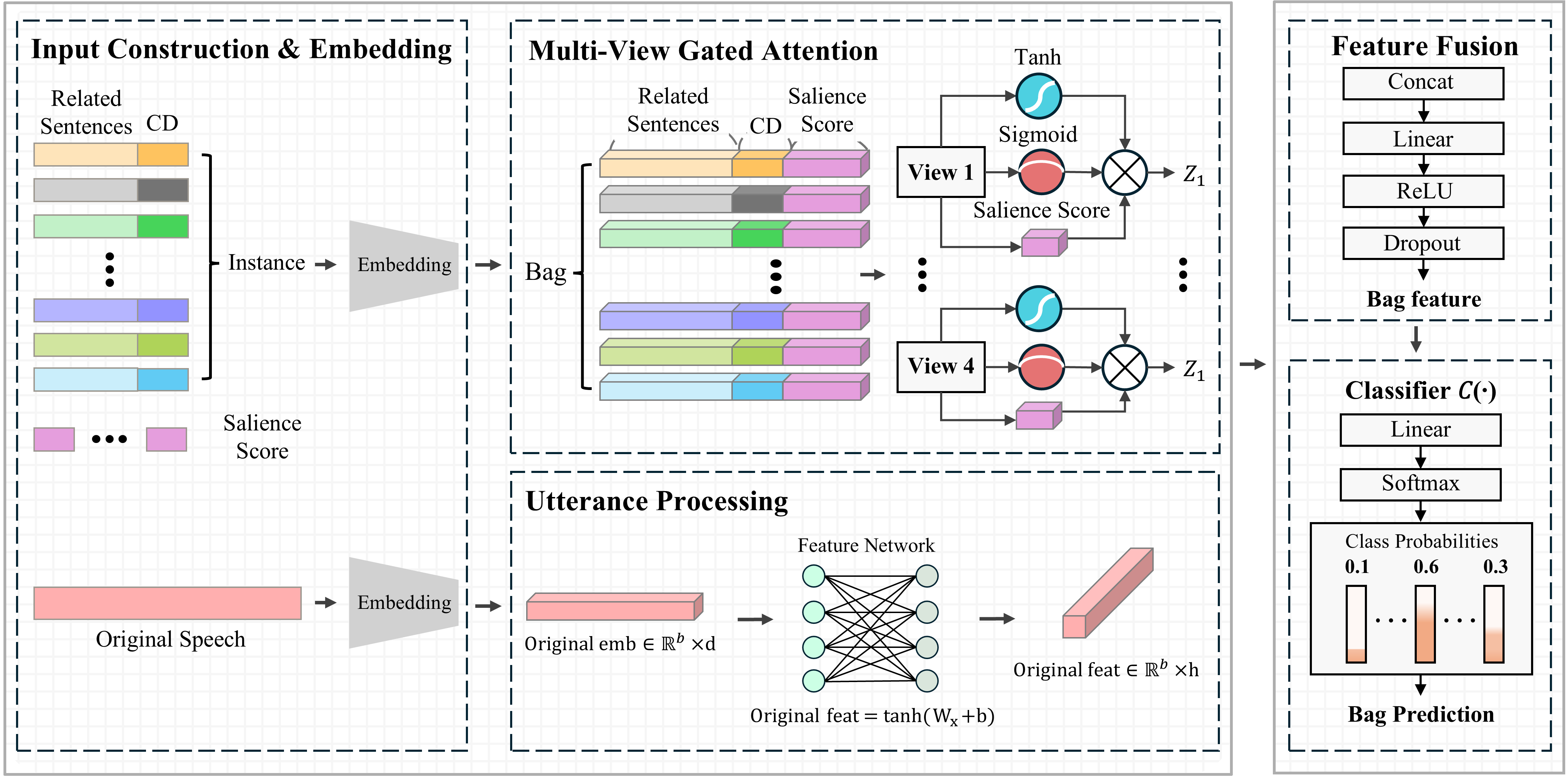}
\caption{Multi-View MIL Architecture for Cognitive Distortion Classification.}
\label{figure3}
\end{figure*}
\label{sec:embedding}

\paragraph{Sentence Embedding} Each bag consists of a sentence embedding vector for the original utterance, which is combined with the instance embeddings. Utterances are embedded as 384-dimensional vectors using the all-MiniLM-L12-v2 model \citep{wang2021}, selected for its favorable balance between representational capacity and computational efficiency. This architecture provides a compact embedding space sufficient for capturing semantic information in short utterances while avoiding unnecessary model complexity.

This embedding serves two purposes: (1) it contributes to bag-level prediction when no instance captures the correct label, and (2) it preserves psychological and contextual information possibly missed at the instance level.

\paragraph{Instance Embedding} Cognitive distortion instances generated from each utterance are also embedded using the same model. Each instance is formed by concatenating the predicted distortion type with the associated sentence and is encoded into a 384-dimensional vector.

For batch processing, bags are padded to the maximum number of instances observed in the dataset. The final input to the MIL model is a sequence combining the sentence embedding and all instance embeddings. Bag labels are one-hot encoded for loss computation only but are not included in the input embeddings.

\subsection{MIL with Multi-View Gated Attention Mechanism}

We propose a Multi-View Gated Attention model for cognitive distortion classification within the MIL framework. Unlike conventional MIL structures, our model integrates LLM-based instance representations and salience scores into the attention mechanism. This architecture processes instances in parallel through multiple independent attention views and subsequently forms a unified representation for bag-level prediction, as illustrated in Figure~\ref{figure3}.

\paragraph{Instance Weighting via Gated Attention}  
To compute the importance of each instance within the bag, we adopt a gated attention mechanism inspired by prior work on attention-based MIL \citep{ilse2018}. The attention score for each instance $i$ is defined as follows:

\begin{equation}
    h_i = \sigma(W_g \cdot x_i) \cdot \tanh(W_f \cdot x_i) \cdot s_i
    \label{eq:3}
\end{equation}

Here, $x_i$ denotes the input vector of instance $i$, $\sigma$ and $\tanh$ represent the sigmoid and tanh activation functions, and $W_g, W_f$ are learnable weight matrices. The sigmoid gate modulates the transformed feature vector from the tanh layer, and the result is scaled by the LLM-derived salience score $s_i$. The final attention score $h_i$ is computed independently for each view.

\paragraph{Multi-View Instance Modeling}  
If attention is computed from a single view, it may fail to capture all relevant instances, as only a subset may be attended to. To address this limitation, we adopt a multi-view attention structure \citep{zhao2017}. Each view independently computes instance-level attention scores using separate gate and feature networks. The final bag-level representation is obtained by averaging the outputs from all $K$ views, as follows:

\begin{equation}
    h_{\text{multi}} = \frac{1}{K} \sum_{k=1}^{K} h^{(k)}
    \label{eq:4}
\end{equation}

Here, $K$ denotes the total number of attention views, and $h^{(k)}$ denotes the aggregated instance representation from the $k^{\text{th}}$ view.

\paragraph{Original Sentence Feature Transformation and Fusion}  
To incorporate global semantic information into the model, the original sentence embedding $z$ is first transformed using a non-linear projection, as follows:

\begin{equation}
    z' = \tanh(W_z \cdot z)
    \label{eq:5}
\end{equation}

The aggregated instance-level representation $h_{\text{multi}}$ from the multi-view attention is then concatenated with the transformed sentence embedding $z'$. This combined vector is passed through a linear projection followed by a ReLU activation to produce the final bag-level representation $v$:

\begin{equation}
    v = \text{ReLU}(W_c \cdot [h_{\text{multi}}, z'])
    \label{eq:6}
\end{equation}

This final representation is used as input to a softmax classifier that predicts the type of cognitive distortion. Together, Equations~\ref{eq:3}--\ref{eq:6} define the core components of our model, which integrate instance-level attention with global sentence-level context.

\subsection{Prediction and Evaluation}

The final prediction is obtained by applying a softmax classifier over the fused bag-level representation. Model training is guided by the standard cross-entropy loss for multi-class classification. The learning rate is initialized at 0.0005 and decays linearly by 0.00001 per epoch until reaching a minimum of 0.00001. Early stopping is applied if the validation loss does not improve for 10 consecutive epochs. Detailed model hyperparameters are provided in Table~\ref{table12} in Appendix~\ref{AppendixB}. To evaluate generalization performance, all experiments are repeated with randomized train--validation--test splits. Results are reported as the mean $\pm$ standard deviation of the weighted average F1 score across 10 runs.

To contextualize the performance of our MIL framework, we additionally evaluate three prompt-based LLM baselines—Gemini 2.0 Flash, GPT-4o, and Claude 3.7 Sonnet—using the same ELB-structured input. These models directly predict the cognitive distortion type without instance extraction or MIL aggregation. For a controlled comparison, we apply the same generation parameters as in our main pipeline. Detailed settings are provided in Table~\ref{table11} in Appendix~\ref{AppendixB}.

\section{Experimental Results and Analysis}

\subsection{Effect of ELB on Label Coverage}

\begin{figure}[t]
\centering
\includegraphics[width=\linewidth]{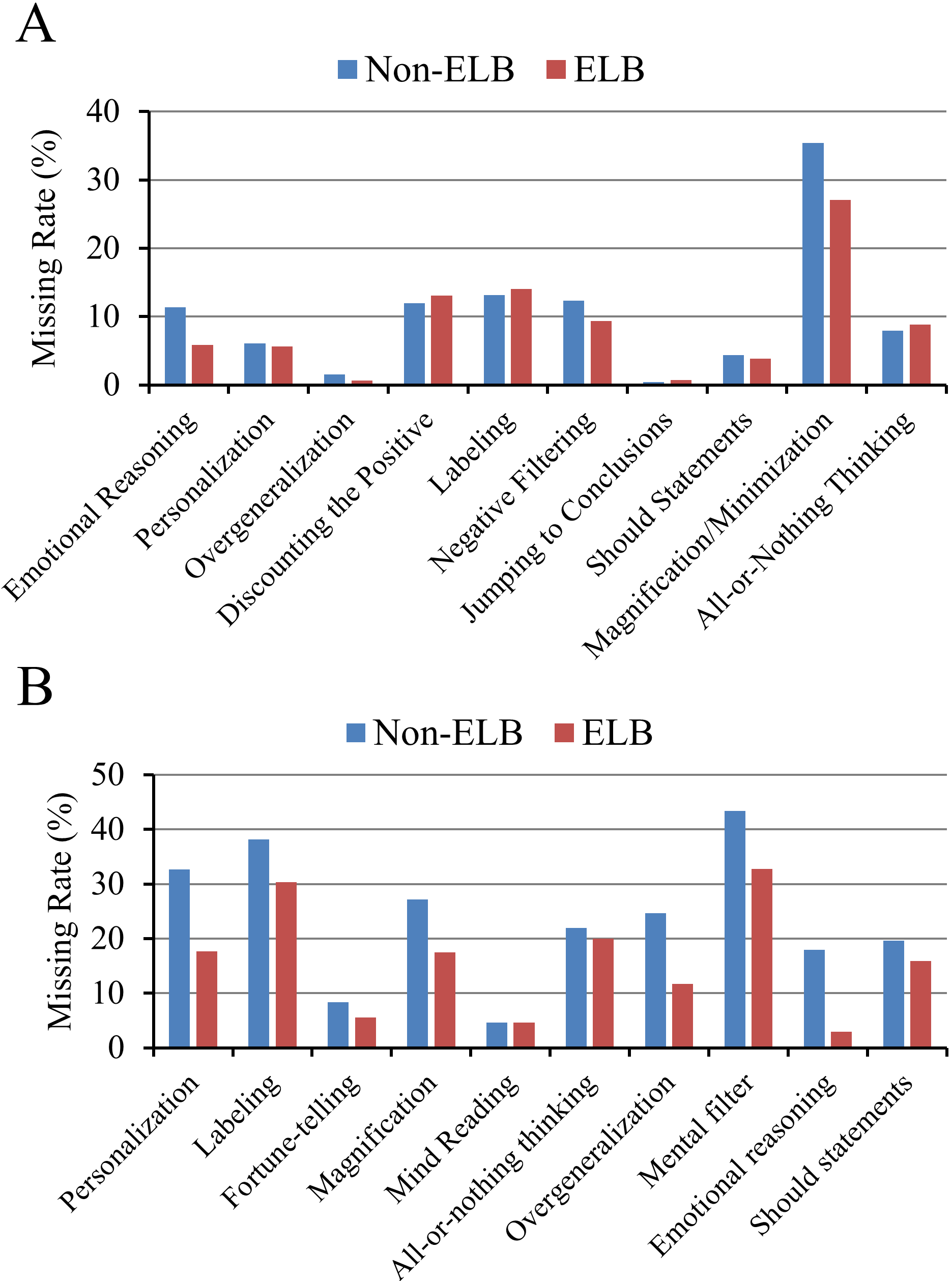}
\caption{Comparative Missing Rates of Cognitive Distortion Instances With and Without ELB (A: KoACD, B: Therapist QA).}
\label{figure4}
\end{figure}

We define a ``missing'' case as one in which none of the instances generated by the LLM includes the ground-truth label of the utterance (bag). As shown in Figure~\ref{figure4}A, incorporating ELB information reduces the overall average missing rate from 10.89\% to 8.93\%, indicating that structuring utterances into ELB components helps the LLM better identify label-relevant expressions. However, while most categories show improvement, types characterized by high semantic ambiguity, such as \textit{Labeling}, continue to exhibit relatively higher omission rates.

As shown in Figure~\ref{figure4}B, the Therapist QA dataset exhibits similar trends, with ELB usage consistently reducing missing rates across all distortion types. This validates the generalizability of ELB structuring for capturing label-relevant expressions across diverse datasets.

\begin{table*}[t]
\centering
\small
\renewcommand{\arraystretch}{1.2}
\setlength{\tabcolsep}{14pt}
\begin{tabular}{lcccc}
\toprule
\multirow{2}{*}[-0.5ex]{\textbf{Methods}} 
& \multicolumn{2}{c}{\textbf{KoACD}} 
& \multicolumn{2}{c}{\textbf{Therapist QA}} \\
\cmidrule(lr){2-3} \cmidrule(lr){4-5}
& \textbf{Val F1} & \textbf{Test F1} 
& \textbf{Val F1} & \textbf{Test F1} \\
\midrule
Baseline & 0.504 $\pm$ 0.019 & 0.473 $\pm$ 0.015 & 0.410 $\pm$ 0.038 & 0.340 $\pm$ 0.037 \\
ELB & 0.519 $\pm$ 0.016 & 0.483 $\pm$ 0.017 & 0.438 $\pm$ 0.028 & 0.378 $\pm$ 0.036 \\
Salience & 0.518 $\pm$ 0.015 & 0.486 $\pm$ 0.014 & 0.428 $\pm$ 0.036 & 0.360 $\pm$ 0.035 \\
ELB + Salience & \textbf{0.529 $\pm$ 0.018} & \textbf{0.505 $\pm$ 0.014} & \textbf{0.460 $\pm$ 0.029} & \textbf{0.394 $\pm$ 0.034} \\
\bottomrule
\end{tabular}
\caption{Effect of ELB and Salience Score Weighting on Classification Performance: All values are 10-run weighted F1 averages; best results are bold.}
\label{table6}
\end{table*}

\setcounter{table}{7}

\begin{table*}[t]
\centering
\small
\renewcommand{\arraystretch}{1.2}
\setlength{\tabcolsep}{8pt}
\begin{tabular}{lcc}
\toprule
\textbf{Cognitive Distortion Type} & \textbf{KoACD (F1)} & \textbf{Therapist QA (F1)} \\
\midrule
Discounting the Positive & \textbf{0.808 $\pm$ 0.036} & -- \\
Should Statements & \textbf{0.852 $\pm$ 0.039} & \textbf{0.460 $\pm$ 0.084} \\
Labeling & \textbf{0.607 $\pm$ 0.052} & \textbf{0.469 $\pm$ 0.102} \\
Mind Reading & -- & \textbf{0.608 $\pm$ 0.061} \\
Personalization & 0.591 $\pm$ 0.036 & 0.409 $\pm$ 0.069 \\
Overgeneralization & 0.440 $\pm$ 0.057 & 0.415 $\pm$ 0.077 \\
Jumping to Conclusions & 0.427 $\pm$ 0.048 & -- \\
Fortune-telling & -- & 0.437 $\pm$ 0.089 \\
All-or-Nothing Thinking & 0.456 $\pm$ 0.077 & 0.164 $\pm$ 0.097 \\
Magnification and Minimization (Magnification in Therapist QA) & 0.390 $\pm$ 0.065 & 0.361 $\pm$ 0.120 \\
Emotional Reasoning & 0.297 $\pm$ 0.052 & 0.164 $\pm$ 0.122 \\
Negative Filtering (Mental filter) & 0.217 $\pm$ 0.047 & 0.214 $\pm$ 0.119 \\
\bottomrule
\end{tabular}
\caption{Per-type F1 scores on KoACD and Therapist QA. The top three scores for each dataset are shown in bold.}
\label{table8}
\end{table*}

\subsection{Input Configuration Comparison}

\setcounter{table}{6}
\begin{table}[t]
\centering
\small
\renewcommand{\arraystretch}{1.2}
\setlength{\tabcolsep}{10pt}
\begin{tabular}{lccc}
\toprule
\textbf{Dataset} & \textbf{Gemini} & \textbf{GPT} & \textbf{Claude} \\
\midrule
KoACD & 0.386 & 0.325 & 0.272 \\
Therapist QA & 0.348 & 0.332 & 0.318 \\
\bottomrule
\end{tabular}
\caption{F1 scores of prompt-based LLM baselines on KoACD and Therapist QA.}
\label{table7}
\end{table}

To analyze the impact of ELB components and LLM-inferred salience scores on cognitive distortion classification, we evaluate four input configurations: a baseline excluding both components, two individual setups incorporating either ELB only or salience scores only, and a final combined configuration including both elements. The results are summarized in Table~\ref{table6} for both the KoACD and Therapist QA datasets. In the KoACD experiment, the configuration using both ELB and LLM-derived salience scores achieves the highest performance, recording a validation F1-score of 0.529 and a test F1-score of 0.505. Notably, the ELB-only configuration outperforms the salience-only variant, suggesting that psychologically grounded structuring provides greater benefits than relying solely on LLM-inferred salience scores.

A similar trend is observed in the Therapist QA dataset, where the combination of ELB and salience scores again yields the best performance. Compared to the DoT-based GPT-4 model with a test F1-score of 0.346 \citep{chen2023}, our model achieves a higher score of 0.394, highlighting the effectiveness of incorporating structured psychological components. Across both datasets, configurations incorporating ELB consistently outperform those relying solely on LLM-derived salience scores, suggesting that structuring utterances into psychological components enhances classification effectiveness.

For reference, we evaluate three ELB-structured prompt-based LLM baselines—Gemini~2.0 Flash, GPT-4o, and Claude~3.7 Sonnet—as reference models. As shown in Table~\ref{table7}, all three models exhibit lower F1 scores than our MIL framework on both datasets, indicating that ELB-augmented zero-shot prediction alone is insufficient for reliable distortion classification. Among the baselines, Gemini achieves the highest performance, followed by GPT and Claude; however, none surpasses our proposed model.

\subsection{Type-wise Performance Analysis}

We evaluate classification performance by distortion type under the ELB + Salience configuration, with the results for both the KoACD and Therapist QA datasets summarized in Table~\ref{table8}.

Across both datasets, \textit{Should Statements} consistently achieves the highest F1-scores. However, a substantial performance gap is observed for this type, with KoACD outperforming Therapist QA.

Expert analysis attributes this discrepancy to linguistic style differences: KoACD contains concise, emotionally direct expressions, whereas Therapist QA includes longer narratives in which obligation-related cues are embedded in complex contexts.

In contrast, distortion types involving emotionally ambiguous or abstract reasoning—such as \textit{Emotional Reasoning} and \textit{Magnification and Minimization}—show lower F1-scores and greater variability, particularly in the English dataset. Further discussion and representative utterances can be found in Table~\ref{table16} in Appendix~\ref{AppendixE}.

\section{Conclusion and Future Work}
We proposed a novel framework that first decomposed utterances into ELB components to extract fine-grained instances, each comprising a predicted distortion type, its associated expression, and a salience score. These components are integrated into a Multiple Instance Learning (MIL) architecture, enabling the model to attend to instances based on semantic relevance and salience. By identifying problematic cognitive patterns from single utterances before extensive dialogue unfolds, this structure facilitates precise early detection and supports timely clinical intervention. While gate scores enable decision traceability to instances, quantitative validation remains future work.
Future work will focus on conducting independent human validation of the extracted ELB components and exploring how extraction errors may propagate to final predictions. We also aim to reduce omission rates for ambiguous distortions and develop quality-aware modeling strategies, as higher instance frequency does not always guarantee superior performance. Finally, applying open-source LLMs to this task is essential to further enhance accessibility and data privacy in sensitive mental health contexts.

\section{Limitations}
In this study, incorporating ELB information into the input configuration significantly reduced the omission rate of correct labels for cognitive distortion expressions. However, several limitations remain. First, true labels are not always captured at the instance level, particularly for distortion types involving subtle emotional cues or subjectively phrased language. This reflects both the inherent representational ambiguity of natural language and the challenges LLMs face in consistently detecting psychologically meaningful patterns across diverse linguistic expressions.

While we did not perform formal human validation of intermediate ELB components or inferred distortion instances, we also did not assess whether the ELB components extracted by the LLM were psychologically valid or aligned with expert interpretations. This introduces a potential limitation: errors in ELB extraction may propagate through the MIL framework, where instance-level representations directly influence bag-level predictions, potentially degrading final classification performance. Addressing this issue through explicit validation of ELB quality remains an important direction for future work.

Moreover, although each cognitive distortion type was equally represented at the utterance level (500 samples per type), the number of inferred instances varied significantly across types. As a result, the MIL model disproportionately focused on distortion types with more abundant or salient instances during attention computation, potentially overlooking valid signals from types with sparser or less prominent instance patterns. These findings underscore the need for more balanced instance generation and attention regulation mechanisms in future work.

Furthermore, although the proposed MIL framework does not generate explicit natural-language explanations, we define interpretability in this study as the ability to relate model predictions to intermediate cognitive distortion instances and their associated text spans. By decomposing an utterance into inferred distortion-level instances, the framework allows us to trace how specific components contribute to the final bag-level classification through gate scores, salience values, and multi-view attention weights.

Nevertheless, we acknowledge that the current approach provides attribution-based interpretability rather than fully validated or clinician-interpretable explanations. In future work, we will aim to develop explanation-generation mechanisms and quantitative validation frameworks grounded in psychological theory to produce more transparent, reliable, and clinically meaningful explanations of model decisions.

Lastly, reliance on proprietary models limits the usability and portability of the framework. In future research, we will evaluate open-source alternatives to improve clinical accessibility and security.

\section{Ethical Considerations}
All data used in this study were obtained from publicly accessible platforms and were anonymized prior to publication. NAVER Knowledge iN and Kaggle public Q\&A data were provided without personal identifiers (e.g., names, contact information, or account details), and no additional identifying information was collected or processed. Therefore, the data adhered to ethical standards of anonymity without requiring further de-identification.

To ensure the validity of the cognitive bias labels, 10 licensed psychological experts independently reviewed and cross-validated 5,000 utterances from the KoACD dataset. Each expert evaluated the appropriateness of the assigned cognitive distortion type based on predefined criteria. All utterances were fully anonymized to prevent any personal identification, and no personal data were collected from the annotators.

All annotators voluntarily participated in the study after being informed of the nature and purpose of the task. They were fairly compensated for their time and expertise, receiving 300{,}000 KRW for annotating 1,000 utterances—an amount calculated to reflect professional hourly rates based on estimated task duration.

The annotation process was conducted under strict anonymity. This study adheres to the principles of responsible research involving human expert participants.

It is important to note that the cognitive distortion detection model proposed in this study is not intended as a substitute for a clinical diagnosis tool. Its use without oversight by qualified mental health professionals is not recommended. As LLM-based models may yield inaccurate or biased interpretation of emotionally sensitive language, any potential real-world applications must involve strict ethical review and expert supervision to ensure safe and responsible deployment.

\section*{Acknowledgments}

This work was supported by the National Research Foundation of Korea (NRF) grant funded by the Korea government (MSIT) (RS-2024-00459226).

\bibliography{custom}

\appendix
\clearpage
\setcounter{table}{8}
\section{Instance Statistics Without ELB Input}
\label{AppendixA}
For completeness, we provide additional statistics summarizing the instance distribution for both datasets when ELB components are not included in the LLM input, as shown in Table~\ref{table9} (KoACD) and Table~\ref{table10} (Therapist QA). These tables serve as a reference for understanding how instance generation differs in the absence of structured psychological input.

\begin{table}[H]
\centering
\small
\renewcommand{\arraystretch}{1.2}
\begin{tabularx}{\linewidth}{l >{\raggedleft\arraybackslash}X}
\toprule
\textbf{Instance Statistics} & \textbf{Value} \\
\midrule
Total Instances & 49,229 \\
Min Instances per Bag & 4 \\
Max Instances per Bag & 19 \\
Avg. Instances per Bag & 10.92 \\
\midrule
\textbf{Cognitive Distortion Type} & \textbf{\# Instances (\%)} \\
\midrule
Jumping to Conclusions & 10,106 (20.6\%) \\
Overgeneralization & 8,809 (17.9\%) \\
Negative Filtering & 5,606 (11.4\%) \\
Personalization & 5,436 (11.0\%) \\
All-or-Nothing Thinking & 4,908 (10.0\%) \\
Emotional Reasoning & 4,728 (9.6\%) \\
Labeling & 3,711 (7.5\%) \\
Should Statements & 2,211 (4.5\%) \\
Magnification and Minimization & 2,169 (4.4\%) \\
Discounting the Positive & 1,545 (3.1\%) \\
\bottomrule
\end{tabularx}
\caption{Instance distribution in KoACD without ELB.}
\label{table9}
\end{table}

\begin{table}[H]
\centering
\small
\renewcommand{\arraystretch}{1.2}
\begin{tabularx}{\linewidth}{l >{\raggedleft\arraybackslash}X}
\toprule
\textbf{Instance Statistics} & \textbf{Value} \\
\midrule
Total Instances & 10,788 \\
Min Instances per Bag & 2 \\
Max Instances per Bag & 21 \\
Avg. Instances per Bag & 6.76 \\
\midrule
\textbf{Cognitive Distortion Type} & \textbf{\# Instances (\%)} \\
\midrule
Emotional reasoning & 2,024 (18.8\%) \\
Mind Reading & 1,528 (14.2\%) \\
Fortune-telling & 1,476 (13.7\%) \\
Overgeneralization & 1,297 (12.0\%) \\
All-or-nothing thinking & 1,015 (9.4\%) \\
Magnification & 938 (8.7\%) \\
Mental filter & 696 (6.5\%) \\
Personalization & 693 (6.4\%) \\
Should statements & 617 (5.6\%) \\
Labeling & 504 (4.7\%) \\
\bottomrule
\end{tabularx}
\caption{Instance distribution in Therapist QA without ELB.}
\label{table10}
\end{table}

\section{Experimental Settings and Model Hyperparameters}
\label{AppendixB}
All experiments were conducted on a local machine equipped with an AMD Ryzen 9 7900X 12-Core Processor (4.70 GHz) without GPU acceleration. The implementation was based on PyTorch 2.6.0.

The hyperparameters used for all LLMs during both ELB extraction and cognitive distortion inference are summarized in Table~\ref{table11}. These values were applied uniformly across GPT-4o, Gemini 2.0 Flash, and Claude 3.7 Sonnet. Specifically, the temperature was set to 1.0 to capture the diversity of valid cognitive distortions that may coexist within a single utterance, a phenomenon recognized in clinical psychological assessment and confirmed by expert psychologists. Although this choice introduces nondeterminism at the instance extraction level, such variability is an intentional design feature rather than noise.

The model hyperparameters for the Multi-View Gated Attention MIL architecture are provided in Table~\ref{table12}. These settings were determined through a combination of grid search and empirical tuning based on validation performance and were consistently used across all experimental configurations.

\begin{table}[H]
\centering
\small
\renewcommand{\arraystretch}{1.2}
\setlength{\tabcolsep}{14pt}
\begin{tabular}{l c}
\toprule
\textbf{Hyperparameter} & \textbf{Value} \\
\midrule
Temperature & 1.0 \\
Top-p & 1.0 \\
Max Tokens & 512 \\
\bottomrule
\end{tabular}
\caption{Hyperparameters for LLMs used in ELB extraction and distortion inference.}
\label{table11}
\end{table}

\begin{table}[H]
\centering
\small
\renewcommand{\arraystretch}{1.2}
\begin{tabular}{l c}
\toprule
\textbf{Hyperparameter} & \textbf{Value} \\
\midrule
Hidden dimension & 128 \\
Dropout rate & 0.3 \\
Number of attention views & 4 \\
Batch size & 32 \\
Number of epochs & 100 \\
Early stopping patience & 10 \\
Optimizer & Adam \\
\bottomrule
\end{tabular}
\caption{Hyperparameters for the Multi-View Gated Attention MIL model.}
\label{table12}
\end{table}

\vspace{2.5em}  

\begin{table*}[t]
\centering
\small
\renewcommand{\arraystretch}{1.2}
\begin{tabularx}{\textwidth}{p{4cm} X X}
\toprule
\textbf{Cognitive Distortion Type} & \textbf{Definition} & \textbf{Examples} \\
\midrule
All-or-Nothing Thinking & Viewing situations in only two categories (e.g., perfect or failure) instead of on a spectrum. & ``If I fail this test, I’m a total failure.'' \\

Overgeneralization & Drawing broad conclusions from a single event or limited evidence. & ``My one friend ignored me, so everyone else will hate me too.'' \\

Mental Filtering & Focusing only on the negative aspects of a situation while ignoring the positive. & ``I only remember my mistake, though I got compliments on my presentation.'' \\

Discounting the Positive & Rejecting positive experiences or compliments by insisting they do not count. & ``People told me I did well, but they were just being polite.'' \\

Jumping to Conclusions & Predicting negative outcomes without evidence. & ``She didn’t text back. She must be mad at me.'' \\

Magnification and Minimization & Exaggerating negative or risky aspects while minimizing positive aspects. & ``One little mistake at work means I’m incompetent.'' \\

Emotional Reasoning & Believing something must be true because you feel it strongly. & ``I feel worthless, so I must be worthless.'' \\

``Should'' Statements & Holding rigid rules about how you or others should behave, leading to guilt or frustration. & ``I should always be productive; otherwise, I’m lazy.'' \\

Labeling & Assigning negative labels to yourself or others based on one event. & ``I made a mistake, so I’m a total failure.'' \\

Personalization & Blaming yourself for events outside your control or assuming excessive responsibility. & ``My friend looks sad; maybe I did something wrong.'' \\
\bottomrule
\end{tabularx}
\caption{Cognitive distortion types in KoACD, adapted from \citep{kim2025}.}
\label{table13}
\end{table*}

\begin{table*}[t]
\centering
\small
\renewcommand{\arraystretch}{1.3}
\begin{tabularx}{\textwidth}{p{3.5cm} X X}
\toprule
\textbf{Cognitive Distortion Type} & \textbf{Definition} & \textbf{Examples} \\
\midrule
Personalization & Personalizing or taking the blame for a situation that, in reality, involves many factors and is outside the person’s control. & ``My son is pretty quiet today. I wonder what I did to upset him.'' \\

Mind Reading & Assuming what others are thinking or inferring the motivations behind their actions. & ``My house was dirty when my friends came over; they must think I’m a slob!'' \\

Overgeneralization & Drawing major conclusions based on limited information. & ``Last time I was in the pool I almost drowned; I am a terrible swimmer and should not go into the water again.'' \\

All-or-Nothing Thinking & Viewing situations as either black or white, with only two possible outcomes. & ``If I cannot get my Ph.D., then I am a total failure.'' \\

Emotional Reasoning & Believing something must be true because you feel it strongly. & ``Even though Steve is here at work late every day, I know I work harder than anyone else at my job.'' \\

Labeling & Assigning a label to oneself or others without sufficient evidence. & ``My daughter would never do anything I disapproved of.'' \\

Magnification & Exaggerating the importance of negative events or outcomes. & ``My professor said he made some corrections on my paper, so I know I’ll probably fail the class.'' \\

Mental Filter & Focusing exclusively on the negative aspects of a situation. & ``My husband says he wishes I was better at housekeeping, so I must be a lousy wife.'' \\

Should Statements & Holding rigid rules about how one or others should behave; these statements may not explicitly include the word `should' or its synonyms (e.g., ought to, must). & ``I should get all A’s to be a good student.'' \\

Fortune-telling & Expecting events to unfold negatively or assuming unfavorable outcomes, even without explicit future tense. & ``I was afraid of job interviews, so I decided to start my own thing.'' \\
\bottomrule
\end{tabularx}

\caption{Cognitive distortion types in Therapist QA, adapted from \citep{shreevastava2021}.}
\label{table14}
\end{table*}

\section{Cognitive Distortion Types}
\label{AppendixC}
This appendix summarizes the definitions and examples of the 10 cognitive distortion types used in each dataset. Although both KoACD and Therapist QA adopt 10-type schemes, the specific categories partially overlap and diverge, reflecting linguistic and cultural differences. These definitions are presented in Table~\ref{table13} for KoACD (adapted from \citep{kim2025}) and in Table~\ref{table14} for Therapist QA (adapted from \citep{shreevastava2021}).

\section{Examples of Cognitive Distortion Instances Inferred by LLMs}
\label{AppendixD}
We present an example of a single utterance (bag) from the KoACD dataset, along with the set of cognitive distortion instances inferred by three LLMs—GPT-4o, Gemini~2.0 Flash, and Claude~3.7 Sonnet. Each instance includes a predicted distortion type, the corresponding text span, and an LLM-assigned salience score, illustrating how a bag is composed of multiple instances in our experimental setting, as shown in Table~\ref{table15}.

\begin{table*}[t]
\centering
\small
\renewcommand{\arraystretch}{2.5}
\renewcommand{\tabularxcolumn}[1]{m{#1}}
\begin{tabularx}{\textwidth}{
    >{\centering\arraybackslash}m{1.2cm} 
    >{\centering\arraybackslash}m{3cm} 
    >{\raggedright\arraybackslash}X 
    >{\centering\arraybackslash}m{1.2cm}}
\hline
\multicolumn{4}{>{\raggedright\arraybackslash}p{\dimexpr\textwidth-2\tabcolsep\relax}}{%
\textbf{Utterance}} \\
\hline
\multicolumn{4}{>{\raggedright\arraybackslash}p{\dimexpr\textwidth-2\tabcolsep\relax}}{%
\textit{I did well in the club interview, but I feel crushed by the thought that I must live up to everyone's expectations. Like the president said, if I don't get selected, does that mean I'm not qualified to be a leader? Do I have to be perfect at everything? I'm definitely lacking. I must do better.}} \\
\multicolumn{4}{>{\raggedright\arraybackslash}p{\dimexpr\textwidth-2\tabcolsep\relax}}{%
\textit{(동아리 면접을 봤다. 발표는 잘했지만, 모두의 기대에 부응해야 한다는 생각에 짓눌린다. 회장 언니 말처럼 뽑히지 못하면 나는 리더가 될 자격이 없는 사람일까? 모든 걸 완벽하게 해내야만 하는 건가? 내가 부족한 건 분명하다. 나는 더 잘해야만 해.)}} \\
\hline
\textbf{LLM} & \textbf{Cognitive Distortion Type} & \textbf{Relevant Sentence (Korean)} & \textbf{Salience score} \\
\hline
 & Should Statements & \textit{I feel crushed by the thought that I must live up to everyone's expectations. ... I must do better.} \textit{(모두의 기대에 부응해야 한다는 생각에 짓눌린다. ... 나는 더 잘해야만 해.)} & 0.444 \\
\cline{2-4}
GPT-4o & Labeling & \textit{If I don't get selected, does that mean I'm not qualified to be a leader? I'm definitely lacking.} \textit{(뽑히지 못하면 나는 리더가 될 자격이 없는 사람일까? 내가 부족한 건 분명하다.)} & 0.333 \\
\cline{2-4}
 & Jumping to Conclusions & \textit{If I don't get selected, does that mean I'm not qualified to be a leader?} \textit{(뽑히지 못하면 나는 리더가 될 자격이 없는 사람일까?)} & 0.222 \\
\hline
 & Jumping to Conclusions & \textit{I'm definitely lacking.} \textit{(내가 부족한 건 분명하다.)} & 0.138 \\
\cline{2-4}
 & Emotional Reasoning & \textit{I feel crushed by the thought that I must meet everyone's expectations.} \textit{(모두의 기대에 부응해야 한다는 생각에 짓눌린다.)} & 0.138 \\
\cline{2-4}
Gemini & Should Statements & \textit{I must meet everyone's expectations. ... I must do better.} \textit{(모두의 기대에 부응해야 한다... 나는 더 잘해야만 해.)} & 0.276 \\
\cline{2-4}
 & All-or-Nothing Thinking & \textit{If I don't get selected, does that mean I'm not qualified to be a leader?} \textit{(뽑히지 못하면 나는 리더가 될 자격이 없는 사람일까?)} & 0.241 \\
\cline{2-4}
 & Labeling & \textit{If I don't get selected, does that mean I'm not qualified? I'm definitely lacking.} \textit{(뽑히지 못하면 나는 리더가 될 자격이 없는 사람일까? 내가 부족한 건 분명하다.)} & 0.207 \\
\hline
 & Discounting the Positive & \textit{I did well in the presentation, but...} \textit{(발표는 잘했지만)} & 0.187 \\
\cline{2-4}
 & Labeling & \textit{I'm definitely lacking.} \textit{(내가 부족한 건 분명하다.)} & 0.172 \\
\cline{2-4}
Claude & Overgeneralization & \textit{If I don't get selected, does that mean I'm not qualified to be a leader?} \textit{(뽑히지 못하면 나는 리더가 될 자격이 없는 사람일까?)} & 0.141 \\
\cline{2-4}
 & Should Statements & \textit{I feel crushed by the thought that I must live up to everyone's expectations. ... I must do better.} \textit{(모두의 기대에 부응해야 한다는 생각에 짓눌린다. ... 나는 더 잘해야만 해.)} & 0.266 \\
\cline{2-4}
 & All-or-Nothing Thinking & \textit{If I don't get selected, does that mean I'm not qualified to be a leader?} \textit{(뽑히지 못하면 나는 리더가 될 자격이 없는 사람일까?)} & 0.234 \\
\hline
\end{tabularx}
\caption{Example of LLM-Inferred Instances from a Single Utterance.}
\label{table15}
\end{table*}

\clearpage
\twocolumn[%
\begingroup
\centering
\small
\renewcommand{\arraystretch}{1.3}
\begin{tabularx}{\textwidth}{>{\raggedright\arraybackslash}X}
\hline
\textbf{Utterance (KoACD dataset)} \\
\hline
\textit{I did well in the club interview, but I feel crushed by the thought that I must live up to everyone's expectations. Like the president said, if I don't get selected, does that mean I'm not qualified to be a leader? Do I have to be perfect at everything? I'm definitely lacking. I must do better.} \\
\textit{(동아리 면접을 봤다. 발표는 잘했지만, 모두의 기대에 부응해야 한다는 생각에 짓눌린다. 회장 언니 말처럼 뽑히지 못하면 나는 리더가 될 자격이 없는 사람일까? 모든 걸 완벽하게 해내야만 하는 건가? 내가 부족한 건 분명하다. 나는 더 잘해야만 해.)} \\
\hline
\textbf{Utterance (Therapist QA dataset)} \\
\hline
\textit{I have been suffering from bulimia for four months now. I realize the health risks and I know I have a problem. I have been trying to stop for a month now with no success. Before this problem I was healthy and now I fear that all my hard work I have completed over the years to be a healthy person are going down the drain. To be honest I am not sure what started my ED, but my main focus is to overcome it. I know that I have some self esteem issues and I will continue to work on that, but do you have any advice or tricks to stop these behaviors that have seemed to become habitual and uncontrollable. I know that getting professional help is probably the best way to go, but that is not me. I have always dealt with my problems in the past and I would like to give this a shot. So if you have any suggestions or tips to help me slowly stop these bulimic behaviors I would appreciate it so much.} \\
\hline
\end{tabularx}
\captionof{table}{Examples of ``Should Statements'' from KoACD and Therapist QA Datasets.}
\label{table16}
\vspace{1em}
\endgroup
]

\section{Cross-Dataset Variation in ``Should Statements''}
\label{AppendixE}
Despite sharing the same distortion label, ``Should Statements'' exhibited a substantial performance gap between the KoACD and Therapist QA datasets. This discrepancy can be attributed to the structural and contextual characteristics of each dataset.
The KoACD dataset, constructed from Korean adolescents' utterances, contains shorter and more direct statements. These utterances are often marked by explicit modal expressions and strong emotional intensity, reflecting adolescents' cognitive style, which tends to favor absolutist reasoning and performance-related anxiety. The utterances are typically centered around school life, peer evaluation, and identity development, with minimal narrative elaboration.
In contrast, the Therapist QA dataset consists of English-language clinical dialogues involving adults. These utterances are generally longer, more reflective, and embedded within broader therapeutic narratives. Statements labeled as ``Should Statements'' frequently appear alongside descriptions of past experiences, diagnoses, and introspective reasoning. This structural complexity introduces interpretive ambiguity, which may challenge classification.
These corpus-level differences in utterance length, discourse structure, and psychological framing help explain the type-level performance variation observed across datasets. Representative examples from each dataset are provided in Table~\ref{table16}.

\section{Prompt Templates}
\label{AppendixF}
We provide the prompt templates used in the LLM-based ELB extraction and cognitive distortion inference processes. The template for ELB extraction is presented in Table~\ref{table17}, and the templates for cognitive distortion instance inference are presented in Tables~\ref{table18} and~\ref{table19}.

\begin{table*}[t]
\centering
\small
\renewcommand{\arraystretch}{1.3}
\begin{tabularx}{\textwidth}{>{\raggedright\arraybackslash}X}
\hline
\textbf{Emotion--Logic--Behavior Extraction} \\
\hline
The user said the following sentence: \\
``\{sentence\}'' \\
\\
Please analyze the sentence according to the following three aspects: \\
\\
1. Emotion: Identify emotional elements or affective states expressed or implied in the sentence, such as anger, sadness, anxiety, frustration, or joy. \\
\\
2. Logic: Identify the reasoning or thought process in the sentence. Look for any assumptions, conclusions, generalizations, or causal relationships. Evaluate whether the reasoning is logical or contains any fallacies. \\
\\
3. Behavior: Identify any behaviors or behavioral intentions mentioned in the sentence. Determine whether the person did something, intends to act, or is avoiding action. \\
\\
For each aspect, provide a brief explanation in the form of a single sentence summarizing the key point. \\
\\
Please respond in the following JSON format: \\
\{ \\
\quad ``emotion'': ``One-sentence summary of the emotional aspect'', \\
\quad ``logic'': ``One-sentence summary of the logical aspect'', \\
\quad ``behavior'': ``One-sentence summary of the behavioral aspect'' \\
\} \\
\\
Note: Each entry should be no more than one sentence. If an aspect is not present, respond with ``Not applicable'' or ``No relevant content found in the sentence.'' \\
\hline
\end{tabularx}
\caption{Prompt for Emotion--Logic--Behavior Extraction.}
\label{table17}
\end{table*}

\begin{table*}[t]
\centering
\small
\renewcommand{\arraystretch}{1.3}
\begin{tabularx}{\textwidth}{>{\raggedright\arraybackslash}X}
\hline
\textbf{Inference of Cognitive Distortion Instances (KoACD)} \\
\hline
The user said the following sentence: \\
``\{sentence\}'' \\
\\
\{emotion\_info\} \\
\{logic\_info\} \\
\{behavior\_info\} \\
\\
Refer to the definitions of the following 10 cognitive distortion types and output all distortion types and salience scores that can be identified in the sentence above. You must select only from the following 10 types: All-or-Nothing Thinking, Overgeneralization, Mental Filter, Discounting the Positive, Jumping to Conclusions, Magnification and Minimization, Emotional Reasoning, Should Statements, Labeling, Personalization. Do not include any other types. \\
\\
Identify all cognitive distortions present in the sentence using the predefined types described in Appendix C, as shown in Table 13. \\
\\
Return all detected cognitive distortions in the following format: \\
{[} \\
\quad \{``type'': ``Cognitive distortion type (must be chosen from the 10 types above)'', ``salience score'': float, ``relevant\_text'': ``Relevant excerpt from the sentence''\}, \\
\quad \{``type'': ``Cognitive distortion type (must be chosen from the 10 types above)'', ``salience score'': float, ``relevant\_text'': ``Relevant excerpt from the sentence''\}, \\
\quad ... \\
{]} \\
\\
Instructions: \\
-- Cognitive distortion types must be written in English only, exactly as listed above (no Korean or parenthetical explanations). \\
\quad Example: ``All-or-Nothing Thinking'' (OK), ``All-or-Nothing Thinking'' (Not OK) \\
-- Choose only from the following 10 types: All-or-Nothing Thinking, Overgeneralization, Mental Filter, Discounting the Positive, Jumping to Conclusions, Magnification and Minimization, Emotional Reasoning, Should Statements, Labeling, Personalization. \\
-- For each distortion, extract a short and relevant excerpt from the sentence that clearly reflects the distortion. \\
\\
Strict output format requirements: \\
1. Return only a JSON array ({[}{]}), with no explanations or additional JSON objects. \\
2. The array must include all detected cognitive distortion objects. \\
3. Each object must contain exactly three fields: ``type'', ``salience score'', and ``relevant\_text''. \\
\hline
\end{tabularx}
\caption{Prompt for Inference of Cognitive Distortion Instances (KoACD).}
\label{table18}
\end{table*}

\begin{table*}[t]
\centering
\small
\renewcommand{\arraystretch}{1.3}
\begin{tabularx}{\textwidth}{>{\raggedright\arraybackslash}X}
\hline
\textbf{Inference of Cognitive Distortion Instances (Therapist QA)} \\
\hline
The user said the following sentence: \\
``\{sentence\}'' \\
\\
\{emotion\_info\} \\
\{logic\_info\} \\
\{behavior\_info\} \\
\\
Refer to the definitions of the following 10 cognitive distortion types and output all distortion types and salience scores that can be identified in the sentence above. You must select only from the following 10 types: All-or-nothing thinking, Overgeneralization, Mental filter, Emotional reasoning, Labeling, Magnification, Should statements, Fortune-telling, Mind Reading, and Personalization. Do not include any other types. \\
\\
Identify all cognitive distortions present in the sentence using the predefined types described in Appendix C, as shown in Table 14. \\
\\
Return all detected cognitive distortions in the following format: \\
{[} \\
\quad \{``type'': ``Cognitive distortion type (must be chosen from the 10 types above)'', ``salience score'': float, ``relevant\_text'': ``Relevant excerpt from the sentence''\}, \\
\quad \{``type'': ``Cognitive distortion type (must be chosen from the 10 types above)'', ``salience score'': float, ``relevant\_text'': ``Relevant excerpt from the sentence''\}, \\
\quad ... \\
{]} \\
\\
Instructions: \\
-- Cognitive distortion types must be written in English name only. \\
\quad Example: ``All-or-nothing thinking'' (Correct), ``All-or-nothing thinking (black and white thinking)'' (Incorrect) \\
-- Cognitive distortion types must be selected ONLY from the 10 types provided: All-or-nothing thinking, Overgeneralization, Mental filter, Emotional reasoning, Labeling, Magnification, Should statements, Fortune-telling, Mind Reading, and Personalization. Do not use any other types. \\
-- For each distortion, extract a short and relevant excerpt from the sentence that clearly reflects the distortion. \\
\\
Strict output format requirements: \\
1. Return only a JSON array ({[}{]}), with no explanations or additional JSON objects. \\
2. The array must include all detected cognitive distortion objects. \\
3. Each object must contain exactly three fields: ``type'', ``salience score'', and ``relevant\_text''. \\
\hline
\end{tabularx}
\caption{Prompt for Inference of Cognitive Distortion Instances (Therapist QA).}
\label{table19}
\end{table*}

\clearpage
\onecolumn
\section{Evaluation Form}
\label{AppendixG}
\begingroup
\centering
\small
\renewcommand{\arraystretch}{1.2}
\begin{tabular}{p{\dimexpr\textwidth-2\tabcolsep\relax}}
\hline
\textbf{Expert Evaluation Form for Cognitive Distortion} \\
\hline
\textbf{Korean} \\[4pt]
안녕하세요, 선생님. \\
본 연구에 참여해주셔서 진심으로 감사드립니다. 이 작업은 청소년의 실제 발화를 기반으로, 해당 발화에 내포된 인지 왜곡 유형을 평가해주시는 작업입니다. 데이터는 파일에 준비되어있습니다. \\[4pt]
각 발화에 대해 아래의 기준에 따라 가장 적절한 인지 왜곡 유형을 선택한 후, 해당 유형에 대한 신뢰도 점수를 1\textasciitilde3점 척도로 평가해 주시면 됩니다. \\[4pt]
1점: 선택된 유형이 문맥에 부합하지 않거나 명확하지 않음 \\
2점: 해당 유형이 어느 정도 나타나지만, 다른 유형과 혼동의 여지가 있음 \\
3점: 해당 인지 왜곡 유형이 명확하고 주된 특징으로 드러남 \\[4pt]
또한, 해당 발화가 다른 인지 왜곡 유형으로도 해석될 가능성이 있다고 판단되시는 경우, 구체적으로 어떤 유형으로도 해석될 수 있는지와 그에 대한 간단한 이유를 함께 적어주시면 감사하겠습니다. 작업 중 불편함이나 감정적 부담을 느끼시는 경우, 언제든지 작업을 중단하거나 일정 조정을 요청하실 수 있습니다. \\[4pt]
문의사항이 있으실 경우 언제든지 연락 주시기 바랍니다. \\[4pt]
감사합니다. \\
\hline
\textbf{English} \\[4pt]
Hello. \\[4pt]
Thank you very much for participating in this study. This task is to evaluate the types of cognitive distortions contained in actual utterances of adolescents. The data is prepared in the file. \\[4pt]
For each utterance, select the most appropriate type of cognitive distortion according to the criteria below, and then evaluate the reliability score on a scale of 1 to 3. \\[4pt]
1 point: The selected type does not fit the context or is unclear. \\
2 points: The type is somewhat present but may be confused with other types. \\
3 points: The cognitive distortion type is clear and stands out as the main feature. \\[4pt]
Additionally, if you believe that the utterance could be interpreted as another type of cognitive distortion, please specify which other types it could be interpreted as and provide a brief explanation of why. \\[4pt]
If you experience any discomfort or emotional distress during the process, you may stop at any time or request a schedule adjustment. \\[4pt]
If you have any questions, please feel free to contact us at any time. \\[4pt]
Thank you. \\
\hline
\end{tabular}
\captionof{table}{Expert Evaluation Form for Cognitive Distortion.}
\label{table20}
\endgroup
\twocolumn

\end{document}